\documentclass{SCIS}
\usepackage{multirow}
\usepackage[table]{xcolor}
\usepackage{threeparttable}
\usepackage{rotating}
\usepackage{overpic}

\usepackage{silence}
\definecolor{mygray1}{gray}{.85}

\begin{document}
%\oa
%%%%%%%%%%%%%%%%%%%%%%%%%%%%%%%%%%%%%%%%%%%%%%%%%%%%%%%
%%% Authors do not modify the information below
\ArticleType{PREPRINT}
% \SpecialTopic{}
\Year{2023}
\Month{}
\Vol{}
\No{}
\DOI{}
\ArtNo{}
\ReceiveDate{}
\ReviseDate{}
\AcceptDate{}
\OnlineDate{}
%%%%%%%%%%%%%%%%%%%%%%%%%%%%%%%%%%%%%%%%%%%%%%%%%%%%%%%

%%% title:
\title{Sora Generates Videos with Stunning Geometrical Consistency}{Sora Generates Videos with Stunning Geometrical Consistency}

% \title{How well Can Sora See the Physical World?\\ A Simple Benchmark  for Video Generative Models}{How well Can Sora See the Physical World: A Simple Benchmark for Video Generative Models}

% \title{Benchmark the Genmetrical Consistency for Video Generative Models}

\author[1]{Xuanyi LI}{}
\author[2]{Daquan ZHOU}{}
\author[2]{Chenxu ZHANG}{}
\author[3]{Shaodong WEI}{} 
\author[1,4]{\\ Qibin HOU}{andrewhoux@gmail.com} 
\author[1,4]{Ming-Ming CHENG}{}

%%% Author information for page head. 
\AuthorMark{Xuanyi LI}

%%% Authors for citation. 
\AuthorCitation{Xuanyi LI, Daquan ZHOU, Chenxu ZHANG, et al}

%%%
\address[1]{VCIP, Nankai University, Tianjin, {\rm 300350}, China}
\address[2]{ByteDance Inc, Singapore}
\address[3]{Wuhan university, Wuhan {\rm 430079}, China}
\address[4]{Nankai International Advanced Research Institute (Shenzhen Futian), Shenzhen, China}

\address[]{\\ xuanyili.edu@gmail.com, zhoudaquan21@gmail.com}
\address[]{Project page: \url{https://sora-geometrical-consistency.github.io/}}

\address[]{First two authors contributed equally.}

\maketitle

% \begin{multicols}{2} 

%%%%%%%%%%%%%%%%%%%%%%%%%%%%%%%%%%%%%%%%%%%%%%%%%%%%%%%
%%% The main text. 
%%%%%%%%%%%%%%%%%%%%%%%%%%%%%%%%%%%%%%%%%%%%%%%%%%%%%%%

\noindent The recently developed Sora model~\cite{videoworldsimulators2024} has exhibited remarkable capabilities in video generation, sparking intense discussions regarding its ability to simulate real-world phenomena. Despite its growing popularity, there is a lack of established metrics to evaluate its fidelity to real-world physics quantitatively. In this paper, we introduce a new benchmark that assesses the quality of the generated videos based on their adherence to real-world physics principles. We employ a method that transforms the generated videos into 3D models, leveraging the premise that the accuracy of 3D reconstruction is heavily contingent on the video quality.
% We use the fidelity of the resulting 3D models as a proxy to gauge how well the generated videos conform to the rules of real-world physics.
From the perspective of 3D reconstruction, we use the fidelity of the geometric constraints satisfied by the constructed 3D models as a proxy to gauge the extent to which the generated videos conform to real-world physics rules.

\begin{figure*}[htp!]
  \centering
  \scriptsize
  \begin{tabular}{cc}
      \includegraphics[width=0.47\textwidth]{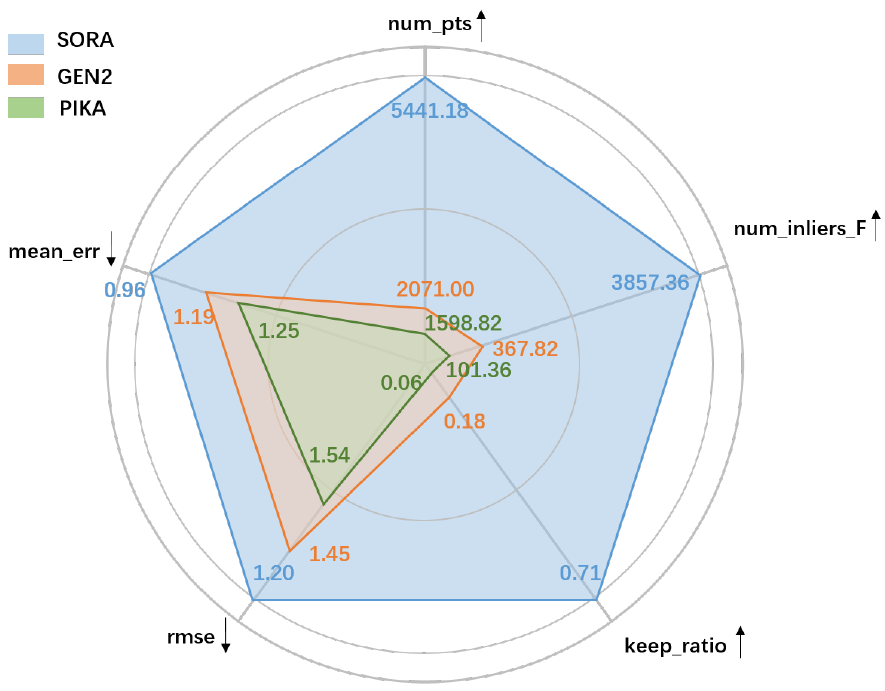} & \includegraphics[width=0.47\textwidth]{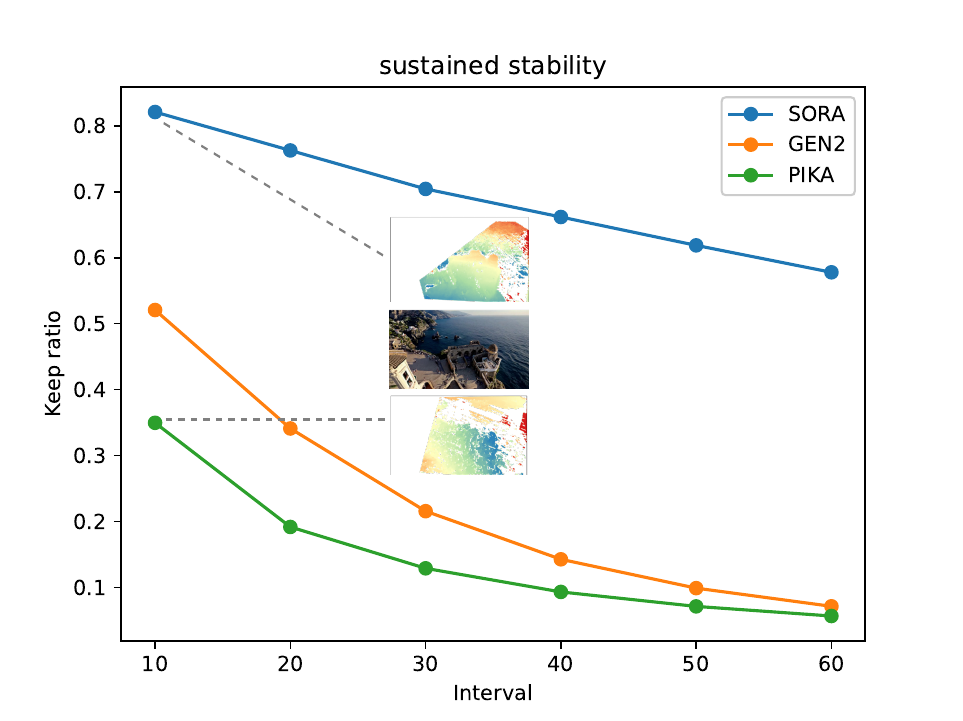} \\
      (a) & (b) \\
  \end{tabular}
  \vspace{-10pt}
  \caption{
   Comparisons among Sora, Pika, and Gen2. (a) shows the quantitative evaluations across five metrics, which we define in Sec.~\ref{sec:method}.
   % \texttt{num\_pts} refers to the total number of initial matching points in the binocular view, and  \texttt{num\_inliers\_F}   refers to the total number of matching points retained after filtering. The \texttt{keep\_ratio} is obtained by the ratio of \texttt{num\_inliers\_F} to \texttt{num\_pts}. 
   % Additionally, for each pair of images, 
%    we calculate the bidirectional geometric reprojection error for $N$ matching points per point of the $F$ matrix, $d(x, x^\prime)$. $x$ and $x'$ are the matching points retained after RANSAC, and  
% $d$ is the distance from a point to its corresponding epipolar line. 
% Finally, we perform an overall statistical analysis of all data to calculate the $\mathrm{RMSE}$ (Root Mean Square Error) and $\mathrm{MAE}$ (Mean Absolute Error). 
For more details, readers can refer to Tab.~\ref{tab:exp1}. 
% performance of different generative video methods under our designed fidelity metric for the Amalfi Coast scene. 
(b) presents the performance of different methods under our designed Sustained Stability metric. For both figures, we can see a significant advantage of Sora over other baselines in terms of geometry consistency.  
   }
  \label{fig:teaser}
\end{figure*}

\section{Introduction} 

Building upon the substantial achievements in image generation, the burgeoning field of text-to-video synthesis (T2V) has been identified as the novel frontier for the application of generative models. Video generation necessitates an evolution over image synthesis due to the intricate requirement of maintaining consistent spatial and temporal relationships across video frames. This intricate process is further compounded by the interpretative challenges associated with brief and often abstract video captions, as well as the limited availability of annotated video-text datasets of high quality.

A new era has dawned in video generation made possible by the advancements in diffusion models, highlighting distinctive frameworks such as Video Diffusion Models~\cite{ho2022video} and Imagen Video~\cite{ho2022imagen}. These pioneering studies have paved the way with innovative conditional sampling methods that facilitate the coherent expansion of videos in both space and time. A noteworthy evolution in this domain is the introduction of MagicVideo~\cite{zhou2022magicvideo}, which has substantially refined the generative process by synthesizing video segments within a compact latent space—a methodology further perfected by Video LDM~\cite{blattmann2023align}. Enhancements continue with the development of MaskDiffusion~\cite{zhou2023maskdiffusion}, which strengthens the generation of text-aligned video content through an improved cross-attention mechanism. Despite these advancements, conventional metrics centered on frame fidelity, motion harmony, and text-video congruence fall short in capturing the geometric quality of the resulting videos.

The recent introduction of the Sora model~\cite{videoworldsimulators2024} has shown itself to be a tour de force in video generation, garnering widespread commendation for its exceptional ability to produce videos with pronounced realism and consistent, logical content along both spatial and temporal vectors. The Sora model's performance is indicative of the significant strides being made in T2V technology, reaffirming the importance of continued innovation in the realm of generative models.
The showcased video clips demonstrate remarkable improvements in quality when compared to the previous leaders in the field, such as SVD~\cite{blattmann2023stable}, Pika Labs~\cite{pikaArt2023}, and Gen-2~\cite{runwayGen2} from Runway.
A standout aspect of Sora is its physical rationality: Despite occasional missteps, many generated clips exhibit adherence to the physical laws and maintain notable geometric properties, which are not adequately captured by previous models.

Furthermore, it is clear that conventional metrics for assessing video generation, such as Fréchet Inception Distance (FID)~\cite{heusel2017gans},  Frechet Video Distance (FVD)~\cite{unterthiner2019fvd}, Inception
Score (IS)~\cite{salimans2016improved}, and Aesthetic Score~\cite{rombach2022high}, etc. do not encompass the dimension of physical accuracy, especially in geometric terms. 
To this end, we consider utilizing the metrics in 3D object quality accessments for the evaluation of the video generation quality. Traditional 3D reconstruction consists primarily of two components: Structure-from-Motion (SFM) and Multi-View Stereo (MVS). SFM, represented by tools like COLMAP~\cite{schonberger2016structure} and OpenMVG~\cite{moulon2017openmvg}, leverages sparse matching and bundle adjustment to estimate camera poses and sparse point clouds. MVS, exemplified by libraries like OpenMVS~\cite{cernea2020openmvs}, takes the results from SFM as input and performs dense matching, mesh reconstruction, and texture mapping to generate 3D models. In recent years, deep learning-based 3D reconstruction methods such as Neural Radiance Fields (Nerf)~\cite{wang2021nerf} and 3D Gaussian Splatting~\cite{kerbl20233d}
have emerged. These methods excel in handling different materials, transparent objects, and underwater environments, resulting in higher rendering quality and enabling fine reconstruction of complex scenes.

Our paper delves into utilizing the quality of 3D object reconstructions derived from Sora's generated videos as a metric to quantitatively evaluate their alignment with physical principles in the context of geometry. 
Specifically, we collect 10 representative video clips generated by Sora that are publicly available on the Internet. 
We then use Pika Labs~\cite{pikaArt2023} and Gen-2~\cite{runwayGen2} to generate videos with the same text prompts. 
Surprisingly, as shown in Fig.~\ref{fig:teaser}, we observe that the videos generated with Sora are good enough for 3D reconstruction, with significant advantages across all selected metrics over the strong baselines.
We hope this simple benchmark could be helpful for video generation models to see how well they can see the physical world.

%%%%%%%%%%%%%%%%%%%%% Table %%%%%%%%%%%%%%%%%%%%%
\begin{table*}[t!]
\centering
\scriptsize
\renewcommand{\arraystretch}{1.2}
\renewcommand{\tabcolsep}{8pt}
\caption{\textbf{Quantitative comparisons on six Sora videos.} The symbols `$\uparrow$/$\downarrow$' indicate that higher/lower scores are better. The best scores are marked in \textbf{bold}. 
%$\Delta$ represents the difference between SAM and the highest score achieved by current cutting-edge COS models.
}
\label{tab:exp1}
\begin{threeparttable}
\begin{tabular}{c|c|c|c|c|c|c}
    \hline
    \rowcolor{mygray1}
    Video name& Method~ 
    &$num\_pts\uparrow$ 
    &$num\_inliers\_F\uparrow$ 
    &$keep\_ratio\uparrow$
    &$mean\_err\downarrow$ 
    &$rmse\downarrow$    \\
    \hline
    \hline
    
    &Sora
    &\textbf{5441.18} &\textbf{3857.36} &\textbf{0.71} &\textbf{0.96} &\textbf{1.20}  \\
    {Amalfi\_coast} 
    &Gen2
    &2071.00 &367.82 &0.18 &1.19 &1.45  \\
    &Pika
    &1598.82 &101.36 & 0.06 &1.25 &1.54  \\  
    \hline
    \hline
    % \multirow{9}{*}
    
    &Sora
    &\textbf{5220.09} &\textbf{3463.91} &\textbf{0.66} &\textbf{0.86} &\textbf{1.12}  \\
    {Art\_museum} 
    &Gen2
    &2480.36 &987.00 &\textbf{0.40} &1.10 &1.35  \\
    &Pika
    &2249.91 &527.73 &\textbf{0.23} &1.27 &1.53 \\   

    \hline
    \hline
    % \multirow{9}{*}
    
    &Sora
    &\textbf{5044.55} &\textbf{3924.09} &\textbf{0.78} &\textbf{0.58} &\textbf{0.80}  \\
    {Big\_sur} 
    &Gen2
    &1899.09 &190.73 &0.10 &1.13 &1.39 \\

    &Pika
    &1662.73 &89.55 &0.05 &1.08 &1.37 \\

    \hline
    \hline

    &Sora
    &\textbf{6198.55} &\textbf{4733.91} &\textbf{0.76} &\textbf{0.89} &\textbf{1.13}  \\
    {Gold\_rush} 
    &Gen2
    &1779.27 &500.36 &0.28 &1.08 &1.34  \\

    &Pika
    &1897.55 &587.73 & 0.30 &1.19 &1.45 \\

    \hline
    \hline

    &Sora
    &\textbf{4911.36} &\textbf{3048.18} &\textbf{0.62} &\textbf{0.92} &\textbf{1.15}  \\
    {Minecraft} 
    &Gen2
    &1445.82 &201.64 &0.14 &1.23 &1.48 \\

    &Pika
    &1196.91 &91.64 &0.08 &0.95 &1.25 \\

    \hline
    \hline

    &Sora
    &\textbf{3790.27} &\textbf{2608.00} &\textbf{0.69} &\textbf{0.99} &\textbf{1.24} \\
    {Santorini} 
    &Gen2
    &1739.91 &365.91 &0.21 &1.26 &1.52  \\

    &Pika
    &1287.18 &60.27 &0.05 &1.07 &1.38 \\

    \hline
    \hline

\end{tabular}
\end{threeparttable}
\end{table*}
%%%%%%%%%%%%%%%%%%%%% Table %%%%%%%%%%%%%%%%%%%%%

\section{Method} 
\label{sec:method}

% Write down a table including the prompts we used to generatio the videos and how we use 3D methods for conversion and the metrics

\paragraph{Process of 3D reconstruction.}
We refrain from modifying the original COLMAP~\cite{schonberger2016structure} and Gaussian Splatting~\cite{kerbl20233d} algorithms to accommodate the characteristics of the generated videos. We utilize Structure-from-Motion (SfM)~\cite{schonberger2016structure} to compute camera poses and then employ Gaussian Splatting for 3D reconstruction.
The detailed metrics used in this benchmark are described in the following.

\paragraph{Metrics design.}
The foundational principle of SFM (Structure from Motion)~\cite{schonberger2016structure} and 3D construction is multi-view geometry, meaning that the quality of the model relies on two main factors: 1) The perspectives of the virtual video's observation cameras must sufficiently meet physical characteristics, such as the pinhole camera; 2) As the video progresses and perspectives change, the rigid parts of the scene must vary in a manner that maintains physical and geometric stability.

Furthermore, the fundamental unit of multi-view geometry is two-view geometry. The higher the physical fidelity of the AI-generated video, the more its two frames should conform to the ideal two–view geometry constraints, such as epipolar geometry.
Specifically, the more ideal the camera imaging of the virtual viewpoints in the sequence video, the more faithfully the physical characteristics of the scene are preserved in the images.
The closer the two frames adhere to ideal two–view geometry, and the smaller the distortion and warping of local features in terms of grayscale and shape, the more matching points can be obtained by the matching algorithm.
Consequently, a higher number of high-quality matching points are retained after RANSAC~\cite{fischler1981random}.
Therefore, we extract two frames at regular intervals from the AI-generated videos, yielding pairs of two-view images.
For each pair, we use a matching algorithm to find corresponding points and employ RANSAC based on the fundamental matrix (epipolar constraint) to eliminate incorrect correspondences.

After elimination, we calculate the average number of correct initial matching points, the average number of retained points, and the average retention ratio. Therefore, we have the following metrics: \texttt{num\_pts} refers to the total number of initial matching points in the binocular view, and  \texttt{num\_inliers\_F}   refers to the total number of matching points retained after filtering. The \texttt{keep\_ratio} is obtained by the ratio of \texttt{num\_inliers\_F} to \texttt{num\_pts}. Additionally, for each pair of images, we calculate the bidirectional geometric reprojection error for $N$ matching points per point of the $F$ matrix, $d(x, x^\prime)$. $x$ and $x'$ are the matching points retained after RANSAC and  
$d$ is the distance from a point to its corresponding epipolar line. Finally, we perform an overall statistical analysis of all data to calculate the $\mathrm{RMSE}$ (Root Mean Square Error) and $\mathrm{MAE}$ (Mean Absolute Error).

%%%%%%%%%%%%%%%%%%%%% Figure %%%%%%%%%%%%%%%%%%%%%
\begin{figure*}[ht!]
  \centering
  % \vspace{-10pt}
  \includegraphics[width=\linewidth]{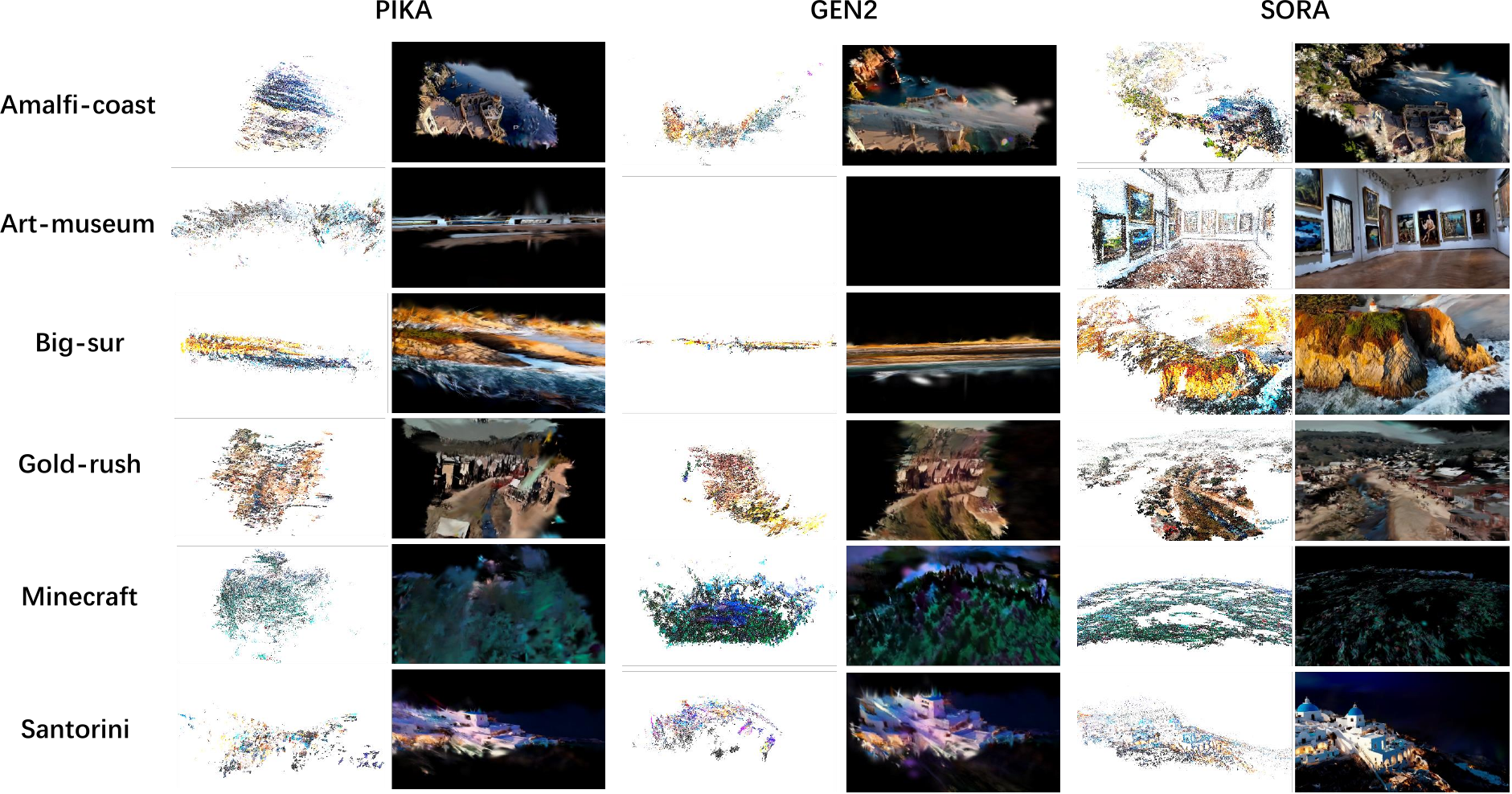}
  \vspace{-15pt}
  \caption{
  Visualizations of point clouds and Gaussian Splatting renderings. This figure presents 3D reconstruction results from videos produced by Pika, Gen2, and Sora. The results of Pika and Gen2 have a limited reconstruction scope with poor geometry and texture. The quality of Sora's reconstructions significantly surpasses that of Pika and Gen2, which can be attributed to two key factors: 1) Sora's ability to produce longer videos, offering more extensive camera information, and 2) the greater frame-to-frame consistency in Sora videos, enabling clearer and more detailed 3D reconstructions. (Note: One of Gen2's videos could not be reconstructed due to insufficient variation in the camera perspective.)
   }
  \label{fig:recon}
\end{figure*}
%%%%%%%%%%%%%%%%%%%%% Figure %%%%%%%%%%%%%%%%%%%%%

\section{Experiments}
In our experiments, a traditional algorithm, SIFT~\cite{ng2003sift}, was chosen for the sparse matching module instead of a more robust deep learning-based matching algorithm. 
This decision aims to prevent the matching performance from being excessively strong, which could potentially mask underlying deficiencies in image quality, such as changes in lighting, texture, and material properties.
Similarly, the dense matching module is implemented with the traditional SGBM algorithm~\cite{hirschmuller2005accurate} for the same reason.
The RANSAC algorithm is employed using the original version available in OpenCV.
We obtain Sora videos directly from the official website. 
To maintain a fair comparison, we utilize the first frame of Sora videos combined with the image2video functionality of both Gen2 and Pika  (using the same prompt) to generate videos with the same scene.

\paragraph{Fidelity metric.}
The total number of frames in each video, with a sampling interval of 30 frames, and a RANSAC threshold of 3 are computed successively starting from the first frame.
The results are shown in Tab.~\ref{tab:exp1}.
The results indicate that while Sora demonstrates similar matching errors compared to the other two methods (still optimal), it acquires several times more correct matching points. 
This suggests that its generated images were the most authentic, with the highest geometrical consistency quality.

% %%%%%%%%%%%%%%%%%%%%% Figure %%%%%%%%%%%%%%%%%%%%%
% \begin{figure*}[htp!]
%   \centering
%   % \begin{overpic}[width=\textwidth]{Imgs/keepratio.pdf}
%   % \end{overpic}
%   % \includegraphics[width=\textwidth]{Imgs/keepratio.pdf}
%   \includegraphics[width=0.8\textwidth]{Imgs/keepratio.pdf}
%   \vspace{-10pt}
%   \caption{
%    Keep ratio metrics under
%    different frame sampling intervals.
%    It can be observed that as the frame interval increases, Sora shows a slow decrease in the preservation ratio of correct matches, while the other two methods exhibit a sharp decrease. This demonstrates the stability and consistency of Sora in preserving physical, imaging, and geometric features over long periods.
%    }
%   \label{fig:keepratio}
% \end{figure*}
% %%%%%%%%%%%%%%%%%%%%% Figure %%%%%%%%%%%%%%%%%%%%%

\paragraph{Sustained stability metric.}
We also compare the variations in the aforementioned $keep\_ratio$ metrics under different frame sampling intervals. This evaluation is referred to as the assessment of sustained stability. 
From the results shown in Fig.~\ref{fig:teaser}(b), it can be observed that as the frame interval increases, Sora shows a slow decrease in the preservation ratio of correct matches, while the other two methods exhibit a sharp decrease.
This demonstrates the stability and consistency of Sora in preserving physical, imaging, and geometric features over long periods.

% % \begin{table*}[t]
% %     \centering
% %     \begin{tabular}{c}
% %          \includegraphics[width=\linewidth]{Imgs/keepratio.pdf}  \\

% %     \end{tabular}
% %     \caption{Caption}
% %     \label{tab:my_label}
% % \end{table*}

% %%%%%%%%%%%%%%%%%%%%% Figure %%%%%%%%%%%%%%%%%%%%%
\begin{figure*}[t!]
  \centering
  \includegraphics[width=\linewidth]{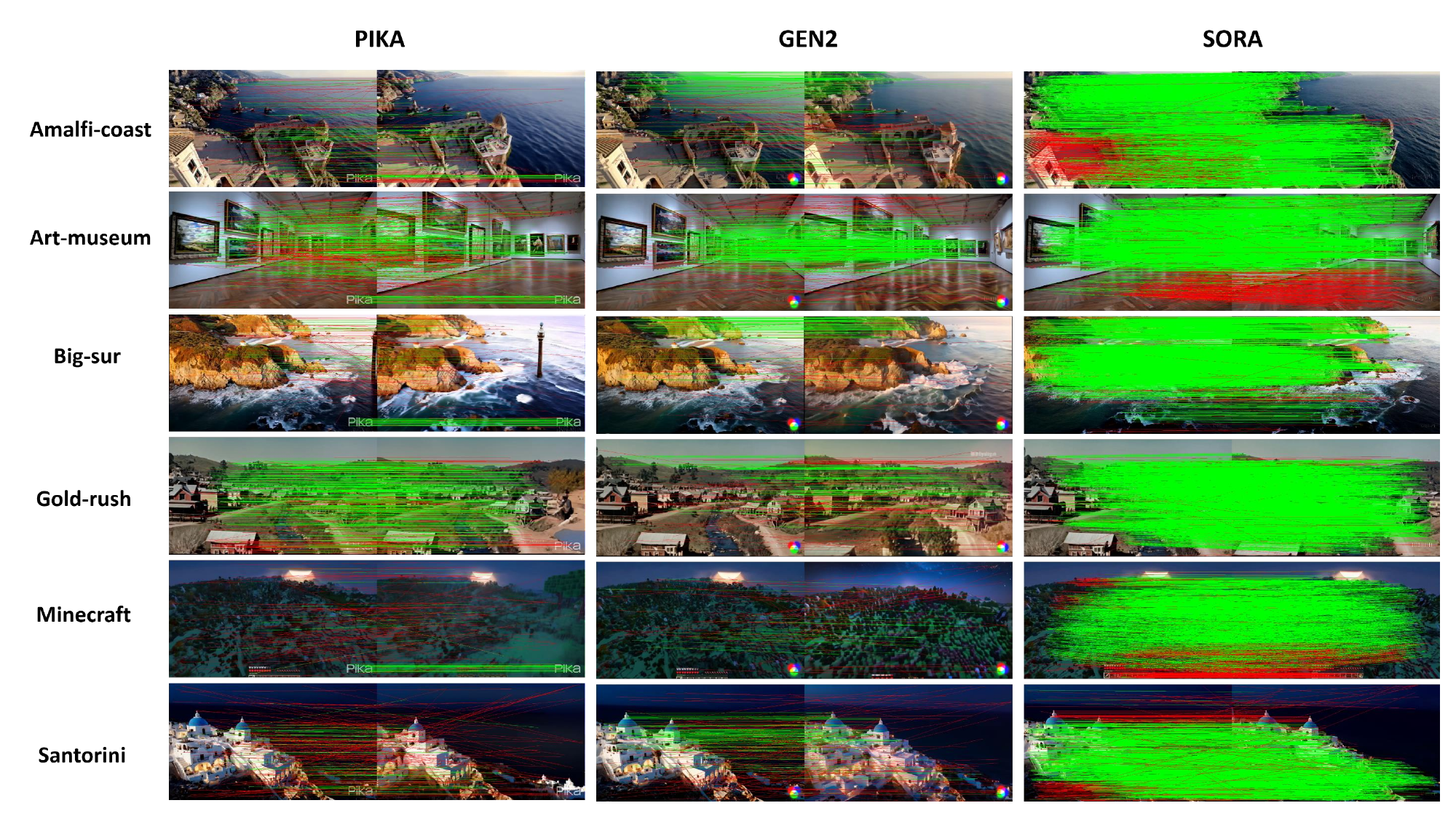}
  \vspace{-20pt}
  \caption{
   Matching result comparisons. In the image, green represents high-quality matching results, while red represents discarded matching results. The presence of more green high-quality matches indicates a higher level of geometric consistency between the two views.
   }
  \label{fig:matching}
\end{figure*}
%%%%%%%%%%%%%%%%%%%%% Figure %%%%%%%%%%%%%%%%%%%%%

\begin{figure*}[t!]
  \centering
  \includegraphics[width=.9\linewidth]{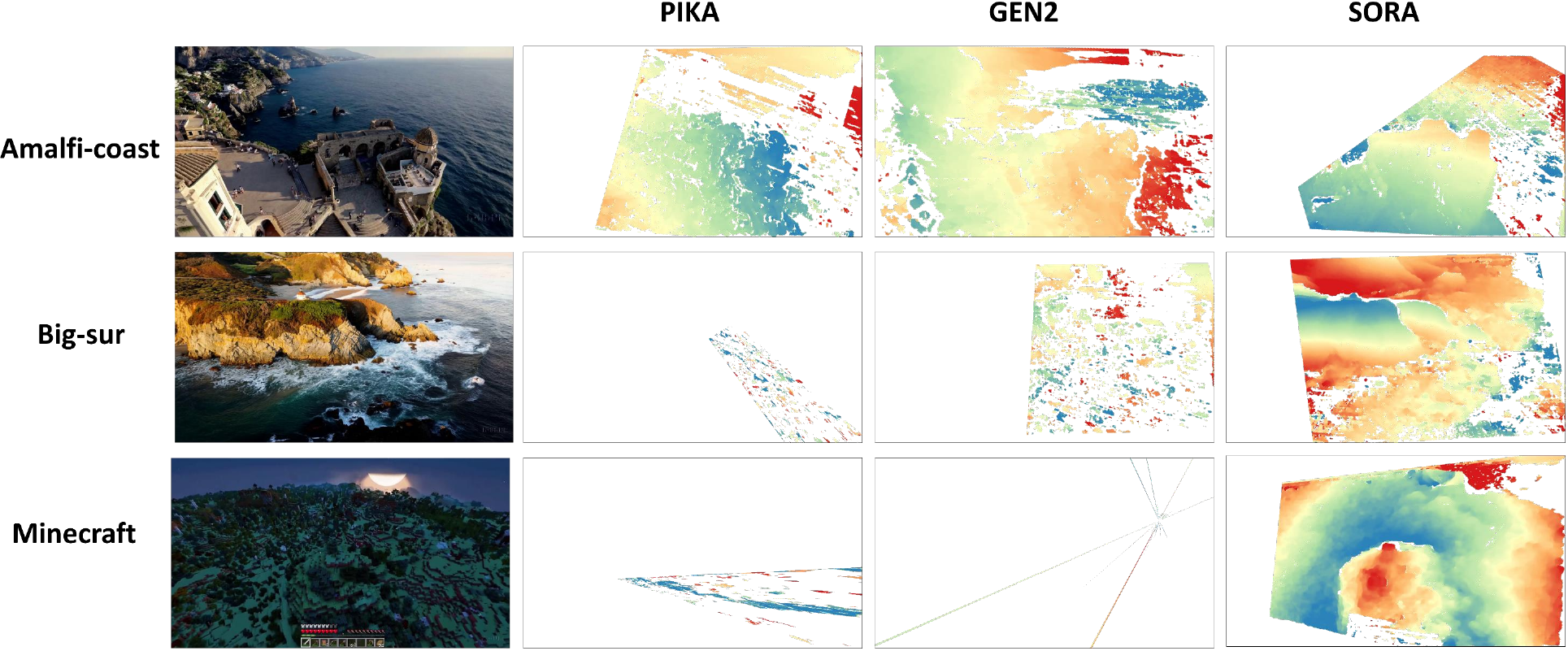}
  \caption{Visualizations of the SGBM stereo matching results reveals that only views strictly adhering to geometric consistency can produce reasonable dense matching results through the SGBM algorithm. It is evident that the videos generated by the Sora method exhibit the most outstanding geometric consistency.
   }
  \label{fig:sgbm}
\end{figure*}

\paragraph{Visualizations.}
First, we demonstrate the 3D reconstruction process using the SFM and Gaussian Splatting methods, showcasing the point clouds and Gaussian Splatting reconstruction results generated by different approaches. 
Fig.~\ref{fig:recon} depicts the 3D reconstruction results from videos produced by Pika, Gen2, and Sora, covering $6$ different scenes. The figure presents both point clouds and Gaussian Splatting visualizations.
Remarkably, the reconstruction quality achieved by Sora is significantly superior to that of Pika and Gen2.
This enhanced performance can be ascribed to two principal reasons: 1) Sora's capability to generate longer videos, which provides a richer set of camera information, and 2) the improved consistency among different frames within Sora-generated videos, which facilitates the reconstruction of clear and intricate 3D geometric structures.
Furthermore, we conduct a visual analysis of the sparse matching results obtained from the videos generated by different methods as shown in Fig.~\ref{fig:matching}.
The videos generated by the Sora method exhibit the highest number of correctly matched points after filtering.
Finally, we feed the rectified stereo images into the SGBM matching algorithm and directly compare the quality of stereo matching results through visualization as shown in Fig.~\ref{fig:sgbm}. 
The visual SGBM stereo-matching results reveal that only views strictly adhering to geometric consistency can produce reasonably dense matching results through the SGBM algorithm.

%%%%%%%%%%%%%%%%%%%%% Figure %%%%%%%%%%%%%%%%%%%%%

\section{Future Work Discussions} The emergence of models like Sora has underscored the need for more precise and holistic assessment tools for video generation tasks. To thoroughly evaluate the quality of generated videos, this research undertakes an initial investigation into the application of 3D reconstruction metrics for examining geometric properties. Beyond geometry, additional physics-based metrics, such as texture authenticity, motion adherence, and interaction logic among scene objects, warrant further consideration. While our primary objective currently is to assess geometric consistency, we acknowledge the importance of these criteria and earmark them for future explorations.

\clearpage

% \newpage

{
\bibliographystyle{IEEEtran}
\bibliography{bibliography}
}

% \end{multicols}
\end{document}